\pdfoutput=1

\documentclass[11pt]{article}

\usepackage{style/acl_arr}
\usepackage{times}
\usepackage{latexsym}

\usepackage[T1]{fontenc}

\usepackage[utf8]{inputenc}
\usepackage{CJKutf8}

\usepackage{microtype}

\usepackage{algorithm}
\usepackage[noend]{algpseudocode}
\usepackage{varwidth}
\usepackage{multirow}
\usepackage{tabularx}
\usepackage{subcaption}
\usepackage{amssymb}
\usepackage{amsmath}
\usepackage{amsthm}
\usepackage{amsfonts}
\usepackage{bbm}
\usepackage{bm}
\usepackage{booktabs}
\usepackage{graphicx}
\usepackage{multirow}
\usepackage{multicol}
\usepackage{enumitem}
\usepackage{mdwlist}
\usepackage{csquotes}

\usepackage[utf8]{inputenc} 
\usepackage[T1]{fontenc}    
\usepackage{hyperref}       
\usepackage{url}            
\usepackage{booktabs}       
\usepackage{amsfonts}       
\usepackage{nicefrac}       
\usepackage{xcolor}         
\usepackage{wrapfig}
\usepackage{makecell}
\usepackage{subcaption}
\usepackage{array}
\usepackage[thinlines]{easytable}
\usepackage{bigstrut}
\usepackage[super]{nth}

\usepackage{enumitem}

\usepackage{geometry}
\usepackage{parskip}
\usepackage{longtable}
\newcounter{myenumi}
\newcounter{myenumii}[myenumi]

\newif\ifcomment\commenttrue

\usepackage[a-1b]{pdfx}

\usepackage{framed}
\usepackage{mdwlist}
\usepackage{latexsym}
\usepackage{colortbl}
\usepackage{xcolor}
\usepackage{nicefrac}
\usepackage{booktabs}
\usepackage{amsfonts}
\usepackage[T1]{fontenc}
\usepackage{bold-extra}
\usepackage{amsmath}
\usepackage{amssymb}
\usepackage{bm}
\usepackage{graphicx}
\usepackage{mathtools}
\usepackage{microtype}
\usepackage{multirow}
\usepackage{multicol}
\usepackage{todonotes}
\usepackage{latexsym,comment}
\usepackage[normalem]{ulem}

\newcommand*{\missingreference}{{\Huge \colorbox{red}{?reference?}}}
\newcommand*{\missingcitation}{{\Huge \colorbox{red}{?citation?}}}
\makeatletter
\def\@setref#1#2#3{%
    \ifx#1\relax
        \protect\G@refundefinedtrue
        \nfss@text{\reset@font\missingreference}%
        \@latex@warning{Reference `#3' on page \thepage \space
            undefined}%
    \else
        \expandafter#2#1\null
    \fi}
\def\@citex[#1]#2{\leavevmode
    \let\@citea\@empty
    \@cite{\@for\@citeb:=#2\do
        {\@citea\def\@citea{,\penalty\@m\ }%
            \edef\@citeb{\expandafter\@firstofone\@citeb\@empty}%
            \if@filesw\immediate\write\@auxout{\string\citation{\@citeb}}\fi
            \@ifundefined{b@\@citeb}{\hbox{\reset@font\missingcitation}%
                \G@refundefinedtrue
                \@latex@warning
                {Citation `\@citeb' on page \thepage \space undefined}}%
            {\@cite@ofmt{\csname b@\@citeb\endcsname}}}}{#1}}
\makeatother

\newcommand{\gem}[1]{\mbox{\textsc{gem}}}
\newcommand{\abr}[1]{\textsc{#1}}

\newcommand{\g}{\, | \,}

\newcommand{\smallemaillink}[2]{{\small \href{mailto://#2}{\texttt{#1}}}}

\newcommand{\hidetext}[1]{}
\newcommand{\ignore}[1]{}

\ifcomment
    \newcommand{\pinaforecomment}[3]{\colorbox{#1}{\parbox{.8\linewidth}{#2: #3}}}
\else
    \newcommand{\pinaforecomment}[3]{}
\fi

\newcommand{\smallurl}[1]{ \begin{tiny}\url{#1}\end{tiny}}

\definecolor{lightblue}{HTML}{3cc7ea}
\definecolor{CUgold}{HTML}{CFB87C}
\definecolor{grey}{rgb}{0.95,0.95,0.95}
\definecolor{ceil}{rgb}{0.57, 0.63, 0.81}
\definecolor{UMDred}{HTML}{ed1c24}
\definecolor{UMDyellow}{HTML}{ffc20e}

\setlist{nosep}

\title{Automatic Song Translation for Tonal Languages}

\author{Fenfei Guo \\
    University of Maryland \\
    \smallemaillink{fguo1@umd.edu} \\\And
    Chen Zhang \\
    Zhejiang University \\
	\smallemaillink{zc99@zju.edu.cn} \\\And
    Zhirui Zhang \\
    Tencent AI Lab\\
	\smallemaillink{zrustc11@gmail.com} \\\AND
	Qixin He \\
	Purdue University\\
	\smallemaillink{heqixin@purdue.edu} \\\And
	Kejun Zhang \\
	Zhejiang University\\
	\smallemaillink{zhangkejun@zju.edu.cn} \\\And
	Jun Xie \\
	Alibaba DAMO Academy\\
	\smallemaillink{qingjing.xj@alibaba-inc.com} \\\AND
    Jordan Boyd-Graber \\
    \abr{cs}, iSchool, \abr{umiacs}, \abr{lsc} \\
    University of Maryland\\
\smallemaillink{jbg@umiacs.umd.edu} \\}

\date{}

\newcommand{\name}{\textsc{g}{\small aga}\textsc{st}}

\begin{document}

\maketitle

\begin{abstract}

This paper develops automatic song translation~(\abr{ast}) for tonal
languages and addresses the unique challenge of aligning words' tones with
melody of a song \emph{in addition to} conveying the original meaning.
We propose three criteria for effective \abr{ast}---preserving
meaning, singability and intelligibility---and design metrics
for these criteria.
We develop a new benchmark for English--Mandarin song translation and
develop an unsupervised \abr{ast}
system, \underline{G}uided \underline{A}li\underline{G}nment
for \underline{A}utomatic \underline{S}ong \underline{T}ranslation
(\name{}), which combines pre-training with three decoding
constraints.
Both automatic and human evaluations show \name{} successfully
balances semantics and singability.

\end{abstract}

\section{Introduction}

\begin{figure}[t]
    \centering
    \includegraphics[width=0.9\linewidth]{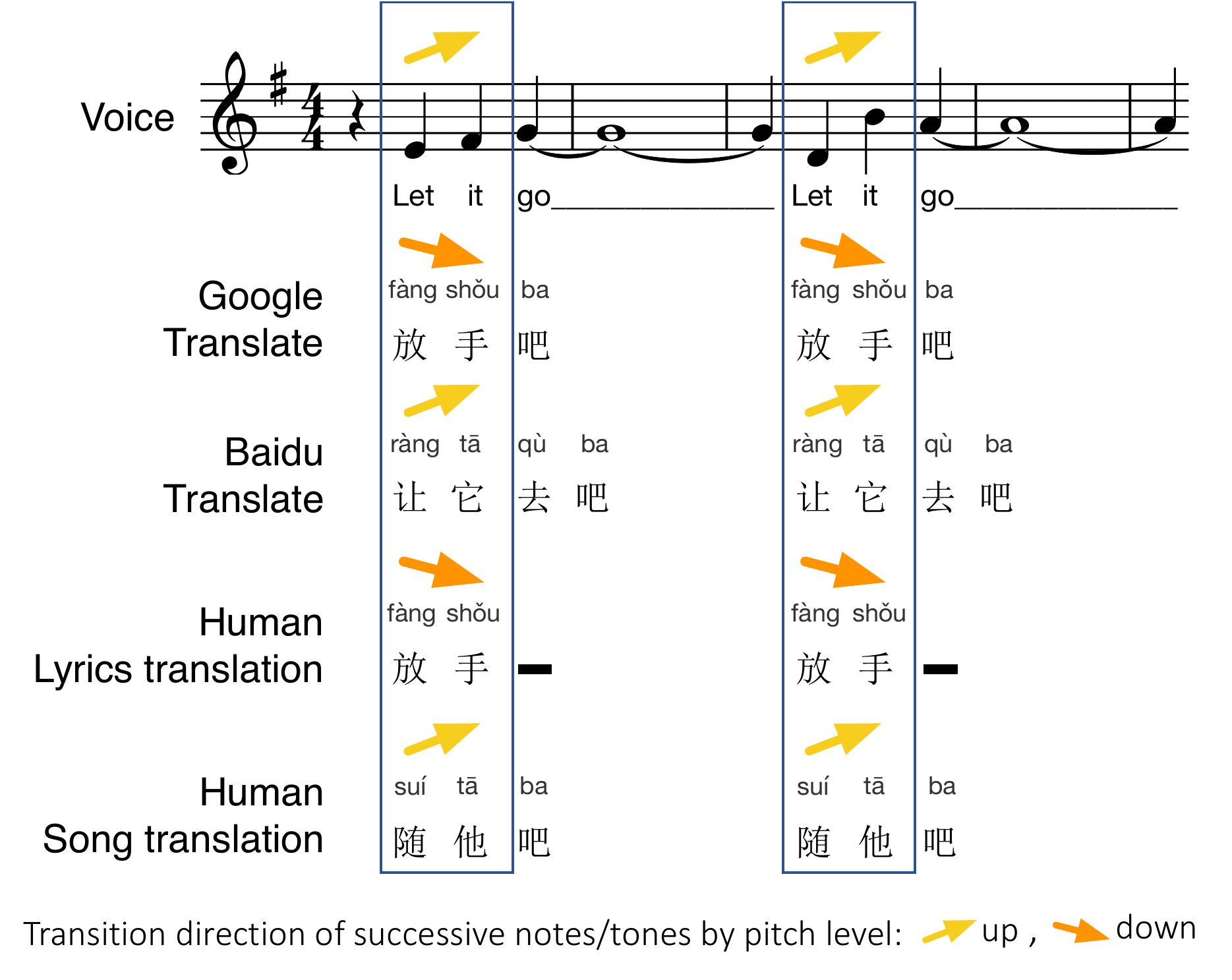}
    \caption{Example Mandarin translations for ``Let it go'' in \textit{Frozen}. Of these, only the official human song translation is something a singer could actually sing: it fits the length of the notes and matches the tones with the pitch of notes.  \name{} finds translations that satisfy these constraints.}
    \label{fig:tonal_align}
\end{figure}

Suppose you are asked to translate the lyrics ``Let it go'' from the
Disney musical \textit{Frozen} into Mandarin Chinese.
Some good, literal translations of this would be A) ``f\`ang
sh\v{o}u'', B) ``f\`ang sh\v{o}u ba'' or C) ``r\`ang t\=a q\`u
ba''~(Figure~\ref{fig:tonal_align}); these get the meaning across and
are the domain of traditional machine translation.
However, what if you needed to sing this song \emph{in Mandarin}?
These literal translations simply do not work: Translations~A and~C do
not match the number of notes and break the original rhythm; while the
tones of Translation~B does not match the pitch flow of the original
melody.

Song translation, unlike translation lyrics for understanding
(subtitling), aims to translate the lyrics so that it can be
\emph{sung} with the original melody.
Therefore, the translated lyrics must match the prosody of the
pre‐existing music in addition to retaining the original meaning.
In \textit{Singable Translations of Songs}, \citet{low2003singable}
says, this is an uncommon and an unusually complex task, a translator
consider rhythm, notes' pitches, phrasing, and stress.
Nonetheless, there are cultural and commercial incentives for more
efficient song translation; \textit{Frozen} alone made over a half a
billion dollars in non-English box office
receipts\footnote{\url{https://www.the-numbers.com/movie/Frozen-(2013)\#tab=international}}
and the musical \textit{Les Mis\'erables} has been performed in over a
dozen languages on stage.

\begin{figure}[t]
    \centering
    \includegraphics[width=0.48\textwidth]{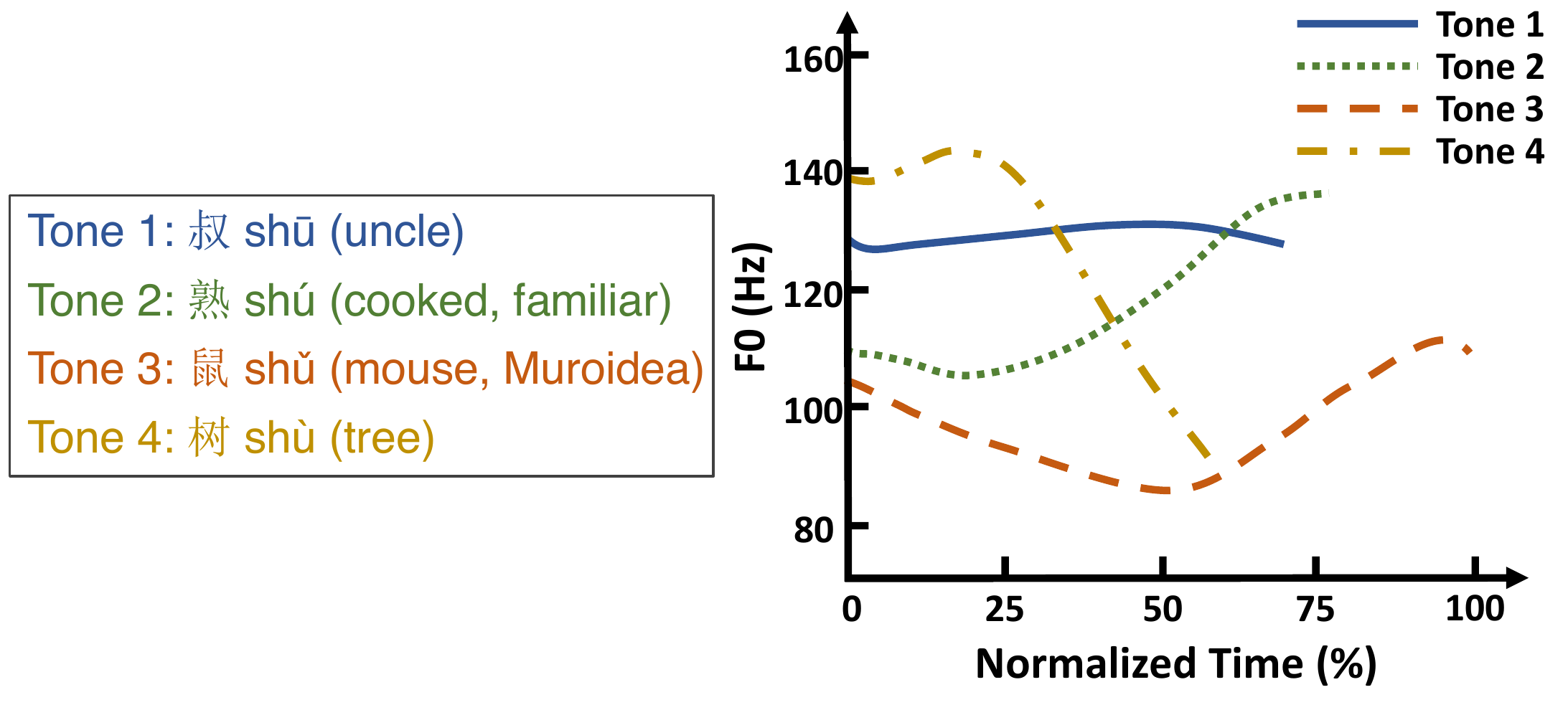}
    \caption{In total languages like Mandarin, the pitch changes the
      meaning of the words~(left).  Each of the four tones in
      Mandarin~(right) has a different pitch profile. Figure
      from \citet{xu1997contextual}.}
    \label{fig:tone_example}
\end{figure}

As we discuss in Section~\ref{sec:background}, while translating
Western songs resembles poetry translation, translating into
\emph{tonal} languages (e.g., Mandarin, Zulu and Vietnamese)
introduces new problems.
In tonal languages, a word's pitch contributes to its meaning
(Figure~\ref{fig:tone_example}); when singing in tonal languages, the
tones of translated words must align with the ``flow'' of the pitches
in the music (Section~\ref{sec:st_tonal}).
For example, if ``f\'ang sh\v{o}u'' were sung instead of ``f\`ang
sh\v{o}u'' (because notes are going up), a listener might hear
``defensive'' instead of the intended meaning.

This paper builds the first system for automatic song
translation~(\abr{ast}) for one tonal language---Mandarin.
Section~\ref{sec:task} proposes three criteria---\textit{preserving
  semantics}, \textit{singability} and
\textit{intelligibility}---needed in an \abr{ast} system.

Guided by those goals, we propose an unsupervised \abr{ast} system,
\underline{G}uided \underline{A}li\underline{G}nment for
\underline{A}utomatic \underline{S}ong \underline{T}ranslation
(\name{}).
\name{} begins with an out-of-domain translation system
(Section~\ref{sec:song_text_train}) and adds song alignment
constraints that favor translations that are the appropriate length
and whose tones match the underlying music
(Section~\ref{sec:method-constraints}).
Naturally, such constraints trade-off between semantic meaning and
singability/intelligibility.
Section~\ref{sec:tradeoff} discusses this trade-off between song
alignment scores and the standard machine translation metric,
\abr{bleu}.

These criteria also form the evaluation for our initial evaluation
(Section~\ref{sec:eval_metrics}).
However, we go beyond an automatic evaluation through a human-centered
evaluation from musicology students.
\name{} creates singable songs that make sense given the original
text, and our proposed alignment scores correlate with human
judgements~(Section~\ref{sec:results}).\footnote{Examples of translated
songs by \name{} at \url{https://gagast.github.io/posts/gagast}.}

\begin{figure}[t]
  \centering
  \includegraphics[width=0.98\linewidth]{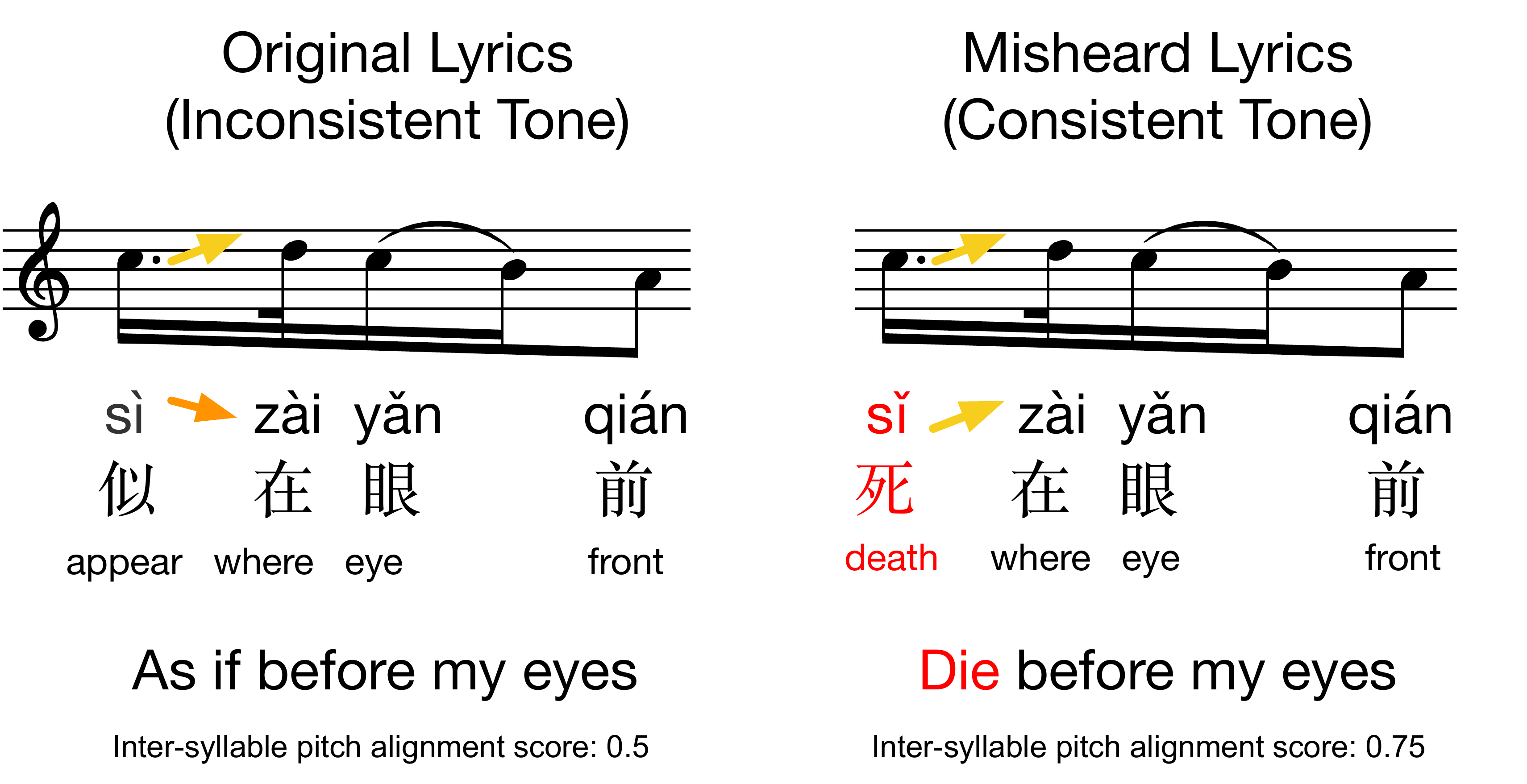}
  \caption{If a song's music doesn't match the tones of the lyrics,
    it can cause the hearer to misunderstand the lyrics.  In this
    example, someone can hear ``s\v{i} z\`ai'' instead of ``s\`i
    z\`ai'', because the notes are going up and ``s\`i z\`ai'' is
    going down. }
  \label{fig:misheard_example}
\end{figure}

\section{Background: Prose, Poetry, and Song Translation}
\label{sec:background}

A spoken language can be divided into two forms: prose, which
corresponds to natural conversation and conventional grammatical
structure; and verse---typically rhythmic and broken into
stanzas--such as poetry and song lyrics.

The vast majority of machine translation research has been focused on
prose translation and has made huge progress; in contrast verse
translation is more difficult as it must obey the rhythmic constraints
and is less developed.
In his \textit{tour de force} work \textit{Le Ton Beau de Marot},
Douglas Hofstadter presents eighty-nine translations of a single poem
to capture the panoply of considerations of what makes the task
difficult~\citep{hofstadter1997ton}.

In western verse, the rhythmic structure are mostly defined by meter,
such as the iambic pentameter for sonnets, which defines the length of
each line, the patterns of long syllables versus short ones and the
stressed ones versus weak ones.
Existing
work~\cite{greene2010automatic,ghazvininejad-etal-2018-neural} use
finite-state constraints to encode both meter and rhyme.
%

Song translation, on the other hand, can be viewed as a translation
where the \emph{melody} defines the constraints.
Reproducing \emph{all} of the essential values of a song---perfectly
matching the meaning, perfectly singable, and perfectly
understandable---is an impossible
ideal~\citep{franzon2008songchoices}.
Thus, tradeoffs are unavoidable.
\citet{low2003singable} argues for prioritizing \textit{singability}
over other qualities such as \textit{sense} and \textit{rhyme} since
``effectiveness on stage'' is a practical necessity.
%
Tonal languages (e.g., Mandarin, Zulu and Vietnamese) dramatically
increases the complexity of singability, and introduces a new factor
that could hamper intelligibility.




\subsection{Song Translation for Tonal Languages}
\label{sec:st_tonal}


For tonal languages, pitch contributes to the meaning of words. 
In a conservative estimation, fifty to sixty percent of the world's
languages are tonal~\citep{yip2002tone} and cover over 1.5 billion
people.
For the lyrics to be \textit{intelligible}, the speech tone and music
tone should be correlated~\citep{schneider1961tone}.
If not, the pitch contour could override the intended tone, which
could produce different meanings.
This is not just a theoretical consideration;
Figure~\ref{fig:misheard_example} shows how lyrics can be and have
been misunderstood.\footnote{More examples at \url{https://gagast.github.io/posts/gagast/\#misunderstanding_examples}}

\subsection{Mandarin Tones and how to Sing them}
\label{sec:st_mandarin}

\citet{schellenberg2013realization} summarizes the rules of singing
with tone with a focus on Chinese dialects.  The tonal system of
Mandarin has two components:
\begin{itemize}[leftmargin=*]
\item \textbf{The pitch level and shape of tones.} Four Mandarin tones
  are used since the \nth{19} century (Figure~\ref{fig:tone_example}).
  We denote tones with a diacritic over the vowel whose shape roughly
  matches the shape of the tone.
  The four tones are a high level (tone~1, e.g., sh\={u}o), rising
  (tone~2, y\'{u}), falling-rising (tone~3, w\v{o}) and falling (tone~4, hu\`{a}i).
\item \textbf{The sandhi of tones.} Some combinations of tones have
  difficult articulatory patterns, so words that might normally have
  one tone might take another depending on the context.
  For example ``n\v{i}'' (you) and ``h\v{a}o'' (good) are typically
  both third tone, but when they are together it is pronounced as
  ``n\'i h\v{a}o'' (hello), with the first syllable changing to a
  \emph{second} tone.
  These changes are called
  sandhi~\citep{xu1997contextual,hu2017sungwords}.
\end{itemize}

\begin{figure}[t]
    \centering
    \includegraphics[width=\linewidth]{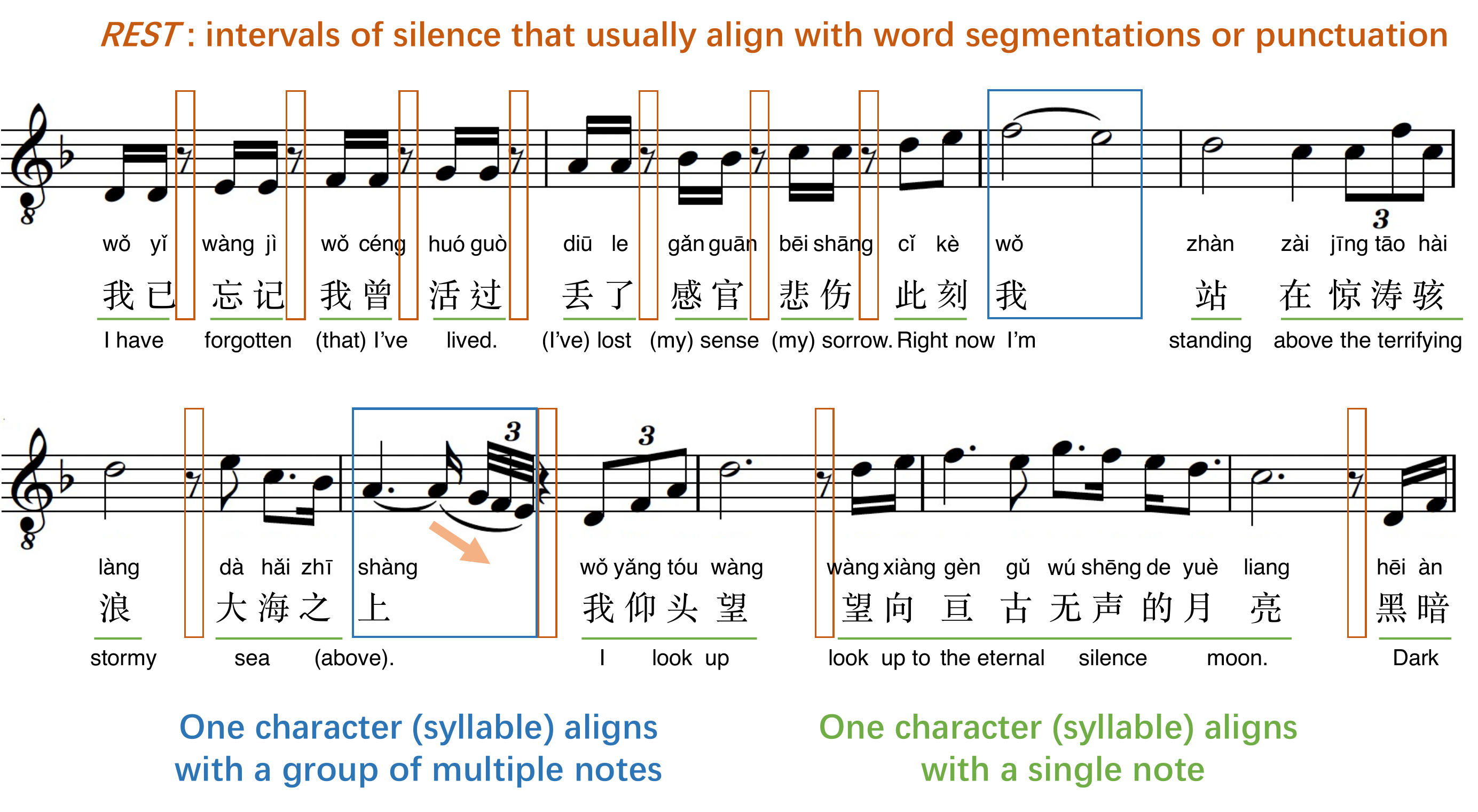}
    \caption{The output of a song translation needs to align syllables
      to the reference melody.  There are several options, as evinced
      by the song ``Love Island~(x\={i}n d\v{a}o)''. \textcolor{orange}{Orange} (top):
      \textit{REST} notes; \textcolor{blue}{Blue} (bottom left): one
      syllable is assigned to a group of multiple notes (which needs
      \textit{tone shape} alignment: the down arrow matches with
      falling tone of ``r\`ang''); \textcolor{green}{Green} (bottom
      right): one syllable is assigned with one note.}
    \label{fig:rest_seg}
\end{figure}

Mandarin tones interact with a sung melody in two
ways~\citep{yang1983language,schellenberg2013realization} to ensure
lyrics are intelligible.
First, at a local level, the \textit{shape of tones} of individual
syllables should be consistent with the musical notes they are matched
with; for example, in ``Love Island''~(Figure~\ref{fig:rest_seg}),
``sh\`ang'' in the blue block has the ``falling'' shape and the group
of notes it assigned to it also falls from an A to a E.
Second, and a global level, the music's \textit{pitch contour} should
align with the tones of the corresponding syllables (taking sandhi
into account).
In practice, we align the transitions between successive syllables and
successive notes (Figure~\ref{fig:tone_contour}) ensuring that the tone matches the relative pitch change~\citep{schellenberg2013realization}.

\section{AST for Tonal Languages}
\label{sec:task}

This section formally defines automatic song
translation~(\abr{ast}) for tonal languages and introduce three
criteria for what makes for a good song translation.
These criteria form the foundation for the quantitative metrics we use
in the experiment.

\subsection{Criteria}
\label{sec:criteria}

There are three criteria that a singable song translation needs to
fulfil.
\paragraph{Preserve meaning.} The translated lyrics should be faithful to the original source lyrics.
%
\paragraph{Singability.} \citet{low2003singable} defines singability as the phonetic compatability of translated lyrics and music.
The translated song needs to be sung without too much difficulty;
difficult consonant clusters, cramming too many syllables into a line, or incompatible tones all impair the singability.
\paragraph{Intelligibility.} The translated song need to be
  understood by the listener.
  This quality has two components.
  First, could a listener produce \emph{any} transcription of the
  lyrics.
  If the lyrics are too fast or garbled because the keywords do not
  fit well with the music, the lyrics are unintelligible.
  Beyond this basic test of recognizability, the lyrics must also be
  accurate: does this transcription match the intended meaning.
  Both aspects matter for a stage performance, since the audience should
  understand the content to follow the plot.
  For pop songs, not understanding all contents could be acceptable
  for some audiences; for example, Adriano Celentano's
  \textit{Prisencolinensinainciusol} sacrifices all intelligibility
  for singability~\cite{bellos-13}.
  However, in more traditional media, hilarious misheard lyrics can
  ruin the audience's experience
  (Figure~\ref{fig:misheard_example}).

\begin{table}[t]
  \small
  \resizebox{\linewidth}{!}{
	\begin{tabular}{ccccccccc}
\toprule
		 {\bf notes} & A3 & C4 & D4 & \textit{REST} & F4 & G4 & F4 & F4 \bigstrut \\ 
		 {\bf pitch level} & 57 & 60 & 62 & & 65 & 67 & 65 & 65 \bigstrut\\
		 {\bf duration} & $\frac{1}{4}$& $\frac{1}{4}$ & $\frac{1}{2}$ & 1 & 3 & $\frac{3}{2}$ & $\frac{1}{2}$ & $\frac{1}{2}$ \bigstrut\\
		 {\bf syllables} & How & a- & bout &  &
                                                        \multicolumn{4}{c}{\cellcolor{blue!25}
                                                        love?} \\
          \bottomrule
        \hline
	\end{tabular} }
	\caption{A snippet of the song ``Seasons of love'' from the
          musical \textit{Rent} that shows the input into
          \name{}. Notes are converted into integer pitches with a
          duration, and syllables are aligned to notes: the ``a'' from
          ``about'' has one note but ``love'' has four.}
	\label{tab:notes_example} 
\end{table}

\subsection{Task Definition}
\label{sec:task_define}

We define the \abr{ast} task as follows: given an aligned pair of
melody~$M$ and source lyrics~$X$, generate translated text~$Y$ in the
target language that aligns with the input melody $M$.
%

Specifically, $X=[x_1, ..., x_L]$ are the input lyrics with~$L$
syllables.
Each syllable~$x_i$ is aligned to a snippet of the melody
(Table~\ref{tab:notes_example}) represented by a sequence of notes.
To represent this to our algorithm, each syllable is aligned to three
components of the melody:
\begin{enumerate}[leftmargin=*]
  \item A sequence of pitch values $\mathbf{p}_i \equiv [p_i^0,
    \dots]$ with $|\mathbf{p}_i|\ge 1$ where an integer
    value of 1.0 means a semitone (e.g., between C and
    C-sharp).
  \item The duration of those notes $\mathbf{d}_i \equiv [d_i^0,
    ...]$, where 1.0 is a quarter note.  Because it encodes the
    duration of each note, the length of $\mathbf{d}_i$ must be the
    same as the length of $\mathbf{p}_i$.
  \item Sometimes there is a rest (pause) before a lyric is sung.  We
    align this to the following syllable~$i$.  The scalar~$r_i$ is the
    real-valued duration of the \textit{REST} note before note group
    $\mathbf{p}_i$. If no \textit{REST} exists before $\mathbf{p}_i$,
    $r_i = 0.0$.
\end{enumerate}




\subsection{Constraints for Aligning Lyrics to Music}
\label{sec:constraints}

To make translated songs singable and intelligible, we summarize three
desirable properties of that the \abr{ast} lyric outputs should have
if they are to match the underlying melody.
Each of these induces a score function which we will use both in our
objective functions for constrained translation and for our evaluation
metrics.

\subsubsection{Length Alignment}

The number of syllables~$L_y$ in translated lyrics~$Y$ need to match
the number of groups of notes $\mathbf{p}_i$ in the melody $M$, so that
it can be sung with the music.
Within the scope of this paper, we either keep the original grouping in the melody $M$ and have $L_y=L_x$ for reproducing the original music; or strictly produce one target syllable for each single note in the melody.

\subsubsection{Pitch Alignment}
\label{sec:pitch}

For tonal languages, pitch of the music must match the lyrics.
As in Section~\ref{sec:st_mandarin}, there are two types of
pitch alignments: 1) \textit{intra-syllable}, the tone shape of each
syllable~(Figure~\ref{fig:rest_seg} blue box) should align with the
shape of the assigned group of notes; 2) \textit{inter-syllable}, the overall pitch contour of the music phrase should align with the tones of lyrics.
  
\paragraph{Intra-syllable alignment.}
For an \emph{individual} syllable, if it is assigned to more than one
note (e.g., ``love'' in Table~\ref{tab:notes_example}), those notes
must be consistent with the shape of the syllable's
tone~\citep{wee2007unraveling}.
For Mandarin, there are four tones
\citep[Figure~\ref{fig:tone_example}]{xu1997contextual}.
We estimate the shape of the multi-note sequence $\mathbf{p}_i$ by least-square estimation and classify it into one of five categories: level, rising, falling, rising-falling, falling-rising.
%

Specifically, for each group $\mathbf{p}_i$ that $|\mathbf{p}_i| > 1$, we classify it as,
\begin{enumerate*}
\item ``level'', if $p_{max}^i - p_{min}^i \le 1.0$; otherwise, we fit $\mathbf{p}_i$ into $ax^2 + bx + c$ via least-square estimation, and compute the axis of symmetry $l = -b/2a$,
\item ``rising'', if ($l \le p_i^0$ and $a > 0.0$) or ($l \ge p_i^{-1}$ and $a < 0.0$);

\item  ``falling'', if ($l \le p_i^0$ and $a < 0.0$) or ($l \ge p_i^{-1}$ and $a > 0.0$);

\item ``rising-falling'', if $ p_i^{0} < l < p_i^{-1}$ and $a < 0.0$;

\item ``falling-rising'', if $ p_i^{0} < l < p_i^{-1}$ and $a > 0.0$;
\end{enumerate*}

%

We compare the shape with that of syllable $y_i$, and compute
the intra-syllable alignment score $S_{\mbox{intra}}^i$:
\begin{equation}
    S_{\mbox{intra}}^i = \begin{cases}
      1.0 & \text{if the shape matches},  \\
      \epsilon & \text{otherwise},  \\
    \end{cases} 
 \label{eq:shape_score}
\end{equation}
where $\epsilon$ is a small parameter that allows for mismatches.
Of the five patterns, ``level'' can match with any tone, ``rising''
matches with tone~2~(y\'u), ``falling'' matches with tone~4~(hu\`ai),
``falling-rising'' matches with tone~3~(w\v{o}) while
``rising-falling'' matches no Chinese tones.


\begin{figure}[t]
    \centering
    \includegraphics[width=0.95\linewidth]{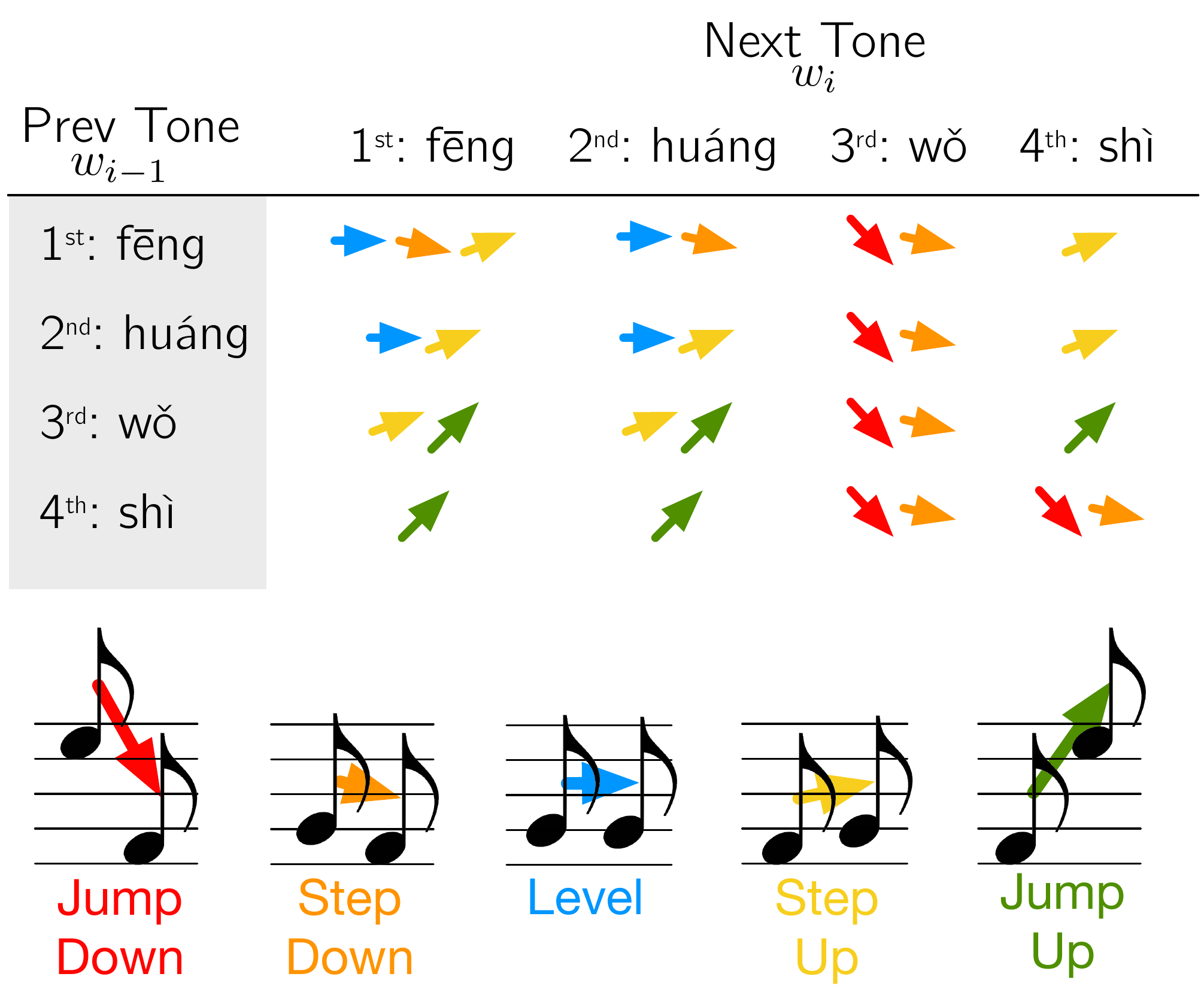}
    \caption{For translated songs in Mandarin to be singable, music
      notes should align the tones of successive characters; this
      becomes our {\bf inter-syllable pitch alignment}. The arrows show
      acceptable transitions in music for two successive
      Mandarin characters ($w_{i-1}, w_{i}$) based on the shape of
      Mandarin tones including sandhi. }
    \label{fig:tone_contour}
\end{figure}

\paragraph{Inter-syllable alignment.}
The second constraint compares the transition directions between consecutive
tones ($t_{i-1}$, $t_i$) of successive syllables ($y_{i-1}$, $y_i$)
that belong to the same word~(see arrows in Figure~\ref{fig:misheard_example}).
These must match the transition directions of music notes
($\mathbf{p}_{i-1}$, $\mathbf{p}_{i}$).\footnote{We compute the directions of two notes group ($\mathbf{p}_{i-1}$, $\mathbf{p}_{i}$) by the first notes ($p_{i-1}^0$, $p_{i}^0$) for simplicity.}
Each transition (the movement from one syllable/note to the
next) can be categorized as \textit{level}, \textit{step up},
\textit{jump up}, \textit{step down} and \textit{jump down}.
We summarize the acceptable transitions for each pair of successive syllables in
Figure~\ref{fig:tone_contour} based on analysis by \citet{yang1983language} and we discuss our choices with more details in Appendix A.2.
Given two syllables ($y_{i-1}$, $y_i$), we compute the local pitch
contour $S_{\mbox{inter}}^i$:
\begin{equation}
    S_{\mbox{inter}}^i = \begin{cases}
      1.0 & \text{if contour matches}, \\
      \epsilon & \text{otherwise}, \\
    \end{cases} 
 \label{eq:contour_score}
\end{equation}
where $\epsilon$ again is a small value to allow mismatches.

\subsubsection{Rhythmic Alignment with Word Segmentation in Mandarin} 
\label{sec:rhythm}



A musical \textit{REST} is a silence separating music.
Recall that in our setup of the data, a scalar $r_i$ denotes if a note
precedes syllable~$i$.
In any language, it is uncommon for a rest to break up a word's syllables.
Thus a good translation should avoid this.  
For Mandarin, creating metrics that capture this are slightly more
complicated because translation systems typically do not explicitly
generate word boundaries.
Thus, we must rely on the output of segmentation systems to know where
word boundaries are.

An exception to this is punctuation (Figure~\ref{fig:rest_seg}).
If a comma, period, or other punctuation is attached to the previous
syllable $y_{i-1}$, then that is a clear signal that it's fine to
pause between them.
Thus, our rest score a syllable~$y_i$ following~$y_{i-1}$ that are
part of different words with probability~$P_{\text{seg}}$\footnote{In practice, we use the cut output by the Jieba toolkit.}, the rest
score is:
\begin{equation}
    S_{R}^i = \begin{cases}
    1.0 & \text{if } r_i > 0.0 \text{ and [punc] after } y_{i-1}, \\
    1.0 & \text{if } r_i = 0.0, \\
    P_{\text{seg}} & \text{if } r_i > 0.0 \text{ and not [punc]}, \\
    \epsilon & \text{otherwise},
    \end{cases}
    \label{eq:rest_constraint}
\end{equation}
where $\epsilon$ is a parameter that represents our tolerance of
having a rest within a word.
   

\begin{figure*}[t]
    \centering
    \includegraphics[width=0.95\textwidth]{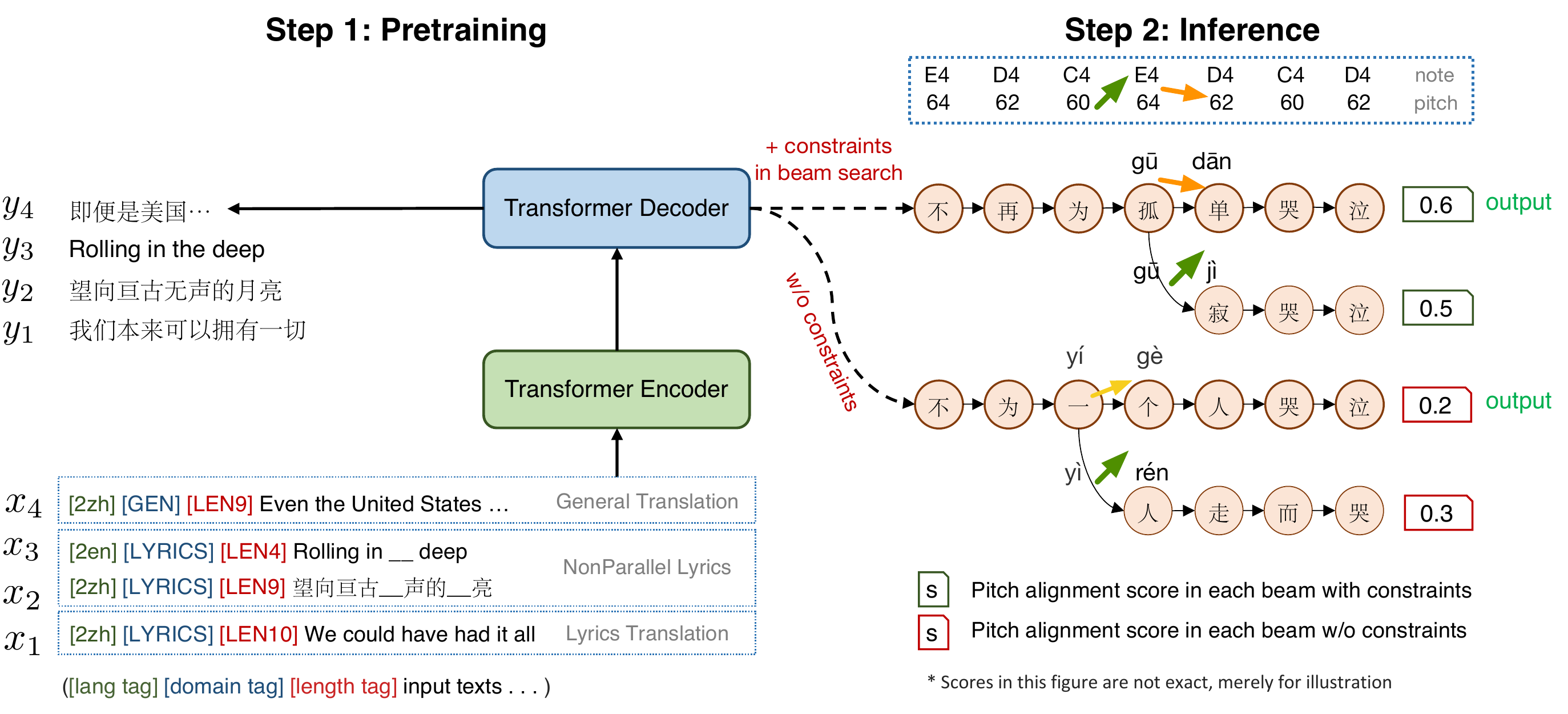}
    \caption{Overview of \name{} for English--Mandarin song
      translation. We first pre-train a lyrics translation model with
      mixture domain data (left); and then add alignment constraints
      in decoding scoring function during inference (right), we use
      unconstrained version as our baseline in the experiment.}
    \label{fig:gagast}
\end{figure*}

\section{GagaST}
\label{sec:gagast}



Ideally, we would build an \abr{ast} system for English--Mandarin song
translation with data-driven models from parallel
data, i.e., aligned triples ($M$, $X$, $Y$).
However, these data are not available in the quantity or quality
necessary for Mandarin: there is not enough data of any quality, and
those that do exist have errors in the syllable-notes alignment.
Thus, we propose an unsupervised \abr{ast} system, \underline{G}uided
\underline{A}li\underline{G}nment for \underline{A}utomatic
\underline{S}ong \underline{T}ranslation (\name{}).
For the pre-training, we collect non-parallel lyrics data in both
English and Mandarin, as well as a small set of lyrics
translation data (Section~\ref{sec:dataset}).



\subsection{Song-Text Style Translation}
\label{sec:song_text_train}

To produce faithful translations in song-text style, we pre-train a
transformer-based translation model with cross-domain data:
translation data in the general domain, the collected monolingual lyrics
data, and a small set of lyric translation data.
%
%
We append domain tags (Figure~\ref{fig:gagast}) before each input
example to control the model to produce translations merely in lyrics
domain during song translation.
For monolingual lyrics data, we adopt
\abr{bart} pre-training~\cite{lewis-etal-2020-bart}.

\subsection{Music Guided Alignment Constraints}
\label{sec:method-constraints}


Without available parallel data to learn the lyric-melody
alignments, we impose
constraints~(Section~\ref{sec:constraints}) in the decoding phase.
Specifically, since all constraints are applied at the
unigram (intra-syllable, \textit{REST}) or bigram~(inter-syllable,
\textit{REST}) level, we apply them at each step of beam search as rewards and penalties in
the scoring function:
\begin{align}
  \log & { P(Y \g X,  M)} =  \sum_{i=0}^{L}\left[ \log{P(y_i \g y_{i-1:0}, X)} \right. \notag \\
    & + \lambda_{\mbox{inter}} \log{S_{\mbox{inter}}^i} + \lambda_{\mbox{intra}} \log{S_{\mbox{intra}}^i} \notag\\
    & + \left. \lambda_{R} \log{S_{R}^i} \right.],
  \label{eq:objective}
\end{align}
where $S_{\mbox{inter}}$, $S_{\mbox{intra}}$, and $S_{R}$ refer to the alignment scores
for inter-syllable pitch alignment, intra-syllable pitch alignment and the rhythm alignment by \textit{REST}.
We introduce three tunable parameters---$\lambda_{\mbox{inter}}$,
$\lambda_{\mbox{intra}}$, and $\lambda_{R}$---that control the importance
of each of the song-specific constraints.

\subsection{Length Control in Pre-training}

To meet the length constraints, we pre-define the syllable-notes
assignments with two strategies:\footnote{A dynamic mapping between
the note sequence and the syllables changes the original rhythm and increases the search space
exponentially.  We leave this to future work.} 1)
\textit{note-to-syllable}, i.e., for each note, we produce one
syllable; 2) \textit{syllable-to-syllable}, we use the original notes
grouping in the input melody, and assign one syllable to each note
group. In this case, the length of target translation is known.
Following \citet{lakew2019controlling}, we use length tag ``[LEN\$i]''
to control the length of outputs during pre-training, where \$i refers
to the length of the target sequence.

\section{Generating Melody-constrained Lyrics and Validating Singability}


This section details data sets, model configuration, and proposed evaluation metrics.
Then we analyze the results and the trade-offs inherent in song
translation.
Our code and data are open-sourced at \url{https://github.com/GagaST}.

\subsection{Training Datasets and Model Configuration}
\label{sec:dataset}

\paragraph{WMT dataset:} news commentary and back-translated news datatsets from \abr{wmt}14 (29.6 million en2zh sentence pairs). No Cantonese texts included and the official Chinese texts can be pronounced in Mandarin by default.


\paragraph{Monolingual lyrics data:}
monolingual lyrics in both Mandarin and English collected from the
web (12.4 million lines of lyrics for Mandarin
and 109.5 million for English after removing duplicates).


\paragraph{Lyrics translation data:}
a small set of lyrics translation data crawled from the
web~\footnote{https://lyricstranslate.com/}
~(140 thousands pairs of
English-to-Mandarin lines). These translations are not singable.

We preprocess all training data with fastBPE~\citep{sennrich-etal-2016-neural}
and a code size of 50,000.
We use encoder-decoder
Transformer~\citep{vaswani2017attention} with 768 hidden units, 12
heads, \abr{gelu} activation, 512 max input
length, 12-12 layers structure (Appendix B for more details).


\subsection{Evaluation Datasets}
\label{sec:eval_data}
For evaluation, we need aligned triples (melody~$M$, source
lyrics~$X$, target reference lyrics~$Y$), where two conditions hold:
1) $M$ and $X$ are syllable-to-note aligned; 2) the reference
$Y$ should be singable and intelligible.
With the confluence of digitization and copyright making such
resources rare, we choose fifty songs from the lyrics translation
dataset that have open-source music sheets on the web and create
aligned triples manually.
However, the reference lyrics in this dataset are not
singable (our primary goal!), we use them to validate that the
translations preserve the original meaning.
Twenty songs comprise the validation set (464 lines) and thirty songs
comprise the test set (713 lines).

\subsection{Evaluation Metrics}
\label{sec:eval_metrics}

An \abr{ast} system for tonal languages should generate translated
songs that are singable and intelligible while conveying the original
meaning.
Evaluating such system is an intrinsically hard
task since all three qualities can be qualitative.
Especially for preserving meaning, the lack of gold references and the
greater tolerance for a loose translation in songs make it difficult to
say how much semantic divergence is acceptable.
Therefore, we first establish evaluations based on the relationship
between lyrics and music and then design human annotations for more
qualitative evaluations.


\subsubsection{Objective Evaluation}

Section~\ref{sec:constraints} outlines three constraints inspired by
music and linguistic theory.
Because these constraints are directly incorporated into the decoding
objective (Equation~\ref{eq:objective}), these will be better than an
unconstrained translation.
However, we want to understand the trade-off between these new
objectives and traditional translation evaluations.

To control for the length of the sentence, we normalize the score to
0--1.0 by the length of alignment pairs $L_i$, that is, based on
Equation~\ref{eq:shape_score},\ref{eq:contour_score} and
\ref{eq:rest_constraint},
\begin{equation}
    s_{[\cdot]} = \sum_1^{L_i}S_{[\cdot]}^i/L_i,
\end{equation}
For the \textbf{length constraint}, we compute: 1) $N_l$, the number of
samples that has length longer than the predefined length $L_i$; 2)
$N_s$, that are shorter than $L_i$. For each case we compute the
average error ratio of $\{\Delta l_i / L_i\}_1^{N_{[\cdot]}}$.
For \textbf{meaning}, although we lack gold singable translations, we
follow the common practice and calculate
\abr{bleu}~\citep{papineni2002bleu} between the translated songs and
the prose translation.

\begin{figure*}[t]
	\centering
	\begin{subfigure}[h]{0.325\textwidth}
	    \captionsetup{justification=centering}
		\centering
		\includegraphics[width=0.9\textwidth]{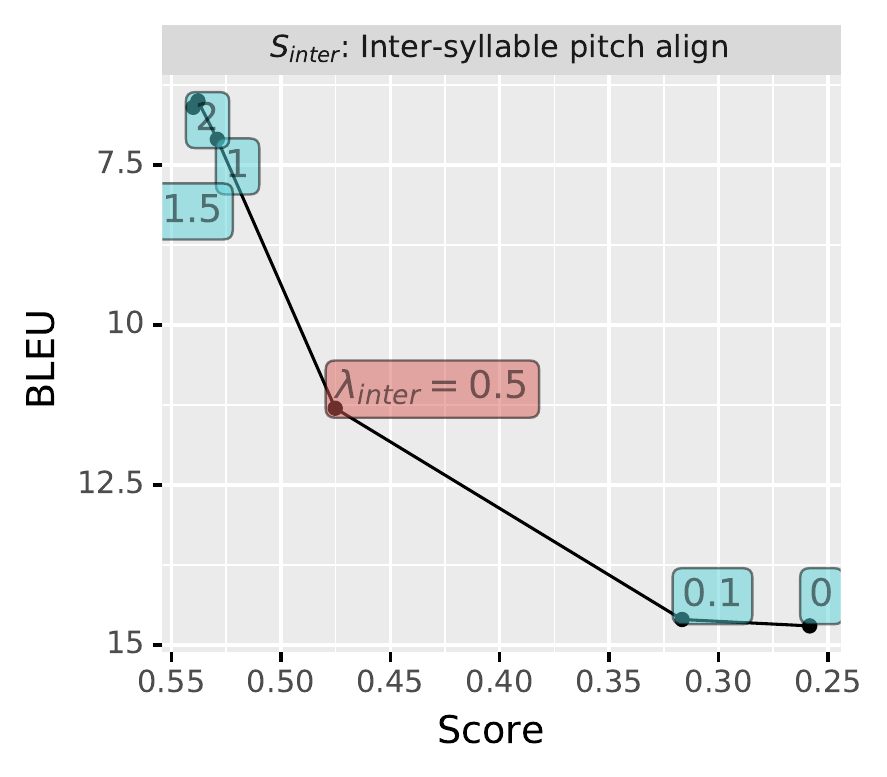}
		\label{fig:constraint_contour}
	\end{subfigure}
	\begin{subfigure}[h]{0.325\textwidth}
	    \captionsetup{justification=centering}
	    \centering
		\includegraphics[width=0.9\textwidth]{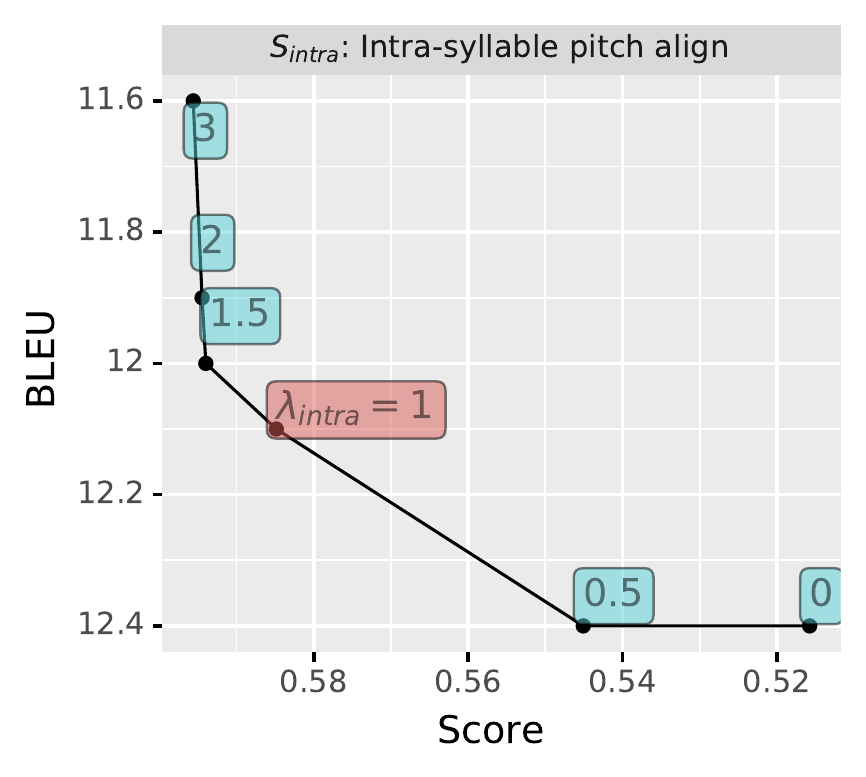}            
		\label{fig:constraint_shape}
	\end{subfigure}
	\begin{subfigure}[h]{0.325\textwidth}
	    \captionsetup{justification=centering}
	    \centering
		\includegraphics[width=0.9\textwidth]{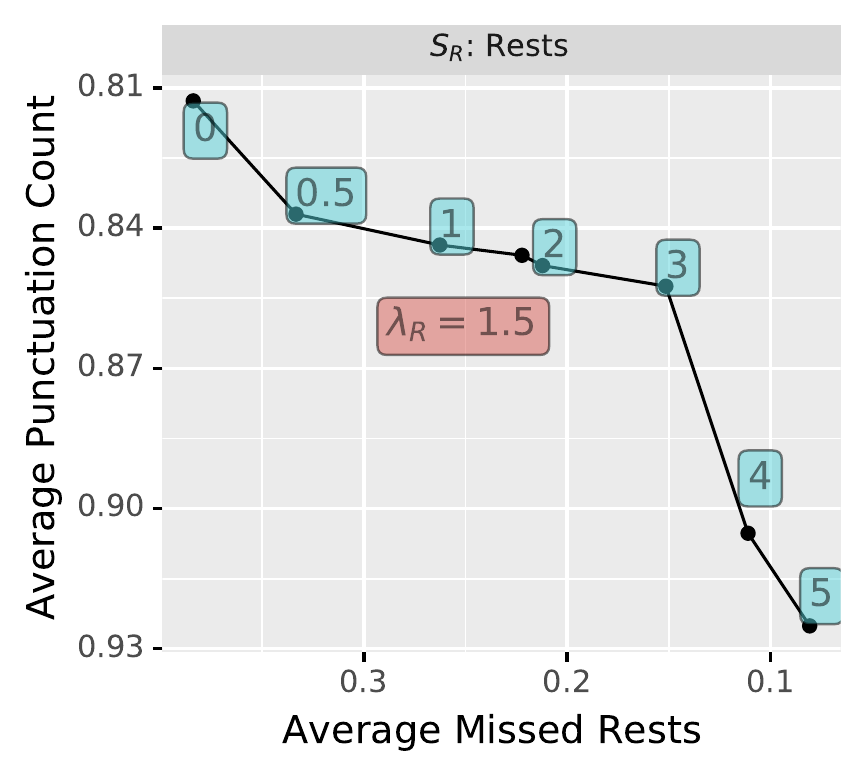}  
		\label{fig:constraint_rest}
	\end{subfigure}
	\caption{Trade-off between meaning ($y$-axis) and
          lyric-music alignments ($x$-axis) while adjusting the tuning
          parameter $\lambda$ on the validation set.  The selected
          value for the tuning parameter $\lambda$ for downstream
          experiments is shown in red (preceeded by $\lambda=$). \abr{rest} constraints do not affect \abr{BLEU}s, but increase the number of [punc]s, which impairs the fluency of the lyrics, thus we select its parameter based on number of [punc]s.}
	\label{fig:ablation_constraints}
\end{figure*}

\begin{table*}[t]
  \small
	\centering
	\resizebox{\linewidth}{!}{
	\begin{tabular}{clcccccc}
		\toprule
		 Syllable-notes & \multicolumn{1}{c}{\multirow{2}{*}{Model}} & \multicolumn{2}{c}{Pitch} & \multicolumn{1}{c}{Rhythm} & \multicolumn{2}{c}{Length} & Meaning\\
		 Assignment & & inter~$\uparrow$ & intra~$\uparrow$ & avg \# of missed rests~$\downarrow$ & longer~$\downarrow$ &  shorter~$\downarrow$ & BLEU~$\uparrow$ \\
		\midrule
		 \multirow{4}{*}{note-to-syllable} & \name{} w/o constraints & 0.28 & - & 0.53 & 9 (0.09) & 0 & 24.0 \\ 
		 & \name{} & 0.51 & - & 0.31 & 26 (0.21) & 0 & 16.9  \\  
		 & \ \ --only inter-syllable & 0.51 & - & 0.45 & 26 (0.21) & 0 & 16.8  \\ 
		 & \ \ --only rest & 0.28 & - & 0.31 & 11 (0.09) & 0 & 23.8  \\ 
        \midrule
        \multirow{5}{*}{syllable-to-syllable} & \name{} w/o constraints & 0.29 & 0.49 & 0.62 & 4 (0.12) & 0 &  22.1  \\
		 & \name{} & 0.50 & 0.55 & 0.28 & 13 (0.13) & 0 & 15.9  \\
		 & \ \ --only inter-syllable & 0.51 & 0.50 & 0.42 & 7 (0.12) & 0 & 15.8 \\
		 & \ \ --only intra-syllable & 0.29 & 0.56 & 0.44 & 4 (0.12) & 0 & 21.6  \\
		 & \ \ --only rest & 0.29 & 0.49 & 0.28 & 5 (0.12) & 0 & 21.6  \\
        \bottomrule
	\end{tabular}
	}
	\caption{Our song-specific constraints with two syllable
          alignment techniques.
          All results here use the same
          pre-training checkpoint and length tags are applied. For
          length score, 9 (0.09) means that 9 out of 713 samples are
          longer than the predefined length with an average ratio
          0.09.
          All constraints have an effect, but inter-syllable pitch alignment has the
          largest.
        }
	\label{tab:main_res} 
\end{table*}

\subsection{Trade-offs between Meaning and Melody-lyric Alignments}
\label{sec:tradeoff}


\name{} adds constraints in the decoding scoring functions to enforce
lyric-music alignments; however, there are trade-offs between
preserving meaning and adhering to these constraints.
To select the importance of these constraints in decoding, we vary the
value of the corresponding parameter $\lambda$
(Equation~\ref{eq:objective}) and analyze how much the \abr{bleu}
score falls on the validation set as we increase the influence of the
parameter.
We set the hyper-parameters where the alignment scores increase fast
while the \abr{bleu} decreases slowly.
The \textit{REST} constraint does not affect the
\abr{bleu}~(Table~\ref{tab:main_res}) but does alter ammount of
punctuation.
Working off the assumption that excessive punctuation is bad, we
select a parameter that minimizes the mismatches between the
\textit{REST} and word boundaries.
We choose (Figure~\ref{fig:ablation_constraints})
$\lambda_{\mbox{inter}}=0.5;\ \lambda_{\mbox{intra}}=1.0;\ \lambda_{R}=1.5$ for all
subsequent experiments.

\subsubsection{BLEU Evaluation}

Table~\ref{tab:main_res} compares \name{} as we ablate constraints
with our two syllable to note alignment strategies
(Section~\ref{sec:gagast}): note-to-syllable and syllable-to-syllable.
As in previous work, the length tag ``[LEN\$i]'' helps lyrics fit the
notes available. In all cases, less than 30 out of 713 lines produces
a longer sentence with ratio less than 0.22; and no short cases.
Thus, because it most closely resembles prior work in controlled
translation and works well in this task, we adopt \name{} with only
length tags and no other constraints as our baseline.
With all of the constraints, \name{} indeed increases both pitch and
rhythm alignments.
It almost doubles the pitch contour alignment score, which affect the
intelligibility the most.

However, these gains come at the cost of \abr{bleu}
score.\footnote{Due the paucity of reliable references, \abr{bleu}
  scores do not correlate with human judgement. For example, three
  official Disney Mandarin song translations have a lower \abr{bleu}
  score (12.3) than our more literal but demonstrably worse automatic
  translations.}
While we believe that the audience would be more accepting of a
less-than-literal translation in a song if it sounds better, we need a
qualitative evaluation to validate that hypothesis.

\subsubsection{Qualitative Evaluation}
\label{sec:human_eval}

The true test of whether \abr{ast} works is whether the songs can be
sung, understood, and enjoyed.
Thus, we follow \citet{sheng2020songmass} and show annotator from a
music school students the resulting sheet music, ask their opinion,
and ask them to sing the songs.
We randomly select five songs from the test set and show the music
sheets~(see Appendix C) of the first ten sentences of each translated
song to five annotators.

Following mean opinion score~\citep[\abr{mos}]{rec1994p} in speech
synthesis, we use five-point Likert scales (1 for bad and 5 for
excellent).
And we evaluate the songs on four dimensions: 1)~\textit{sense},
fidelity to the meaning of the source lyric; 2)~\textit{style},
whether the translated lyric resembles song-text style;
3)~\textit{listenability}, whether the translated lyric sounds
melodious with the given melody; 4)~\textit{intelligibility}, whether
the audience can easily comprehend the translated lyrics if sung with
provided melody.
The last two dimensions require the annotators to
sing the song.

\begin{table}[t]
  \small
	\centering
	\resizebox{\linewidth}{!}{
	\begin{tabular}{cccccc}
		\toprule
		 Model & Song & \textit{sense} & \textit{style} & \textit{listenability} & \textit{intelligibility} \\
		\midrule
		 \multirow{3}{*}{} & Song1 & 3.4 & 3.0 & 3.2 & 3.4 \\
		 & Song2 & 3.6 & 3.9 & 3.4 & 3.8 \\
		 \name{} & Song3 & 3.7 & 3.6 & 3.4 & 3.5 \\
		 w/o constraints & Song4 & 3.2 & 3.0 & 2.8 & 3.0 \\
		 \multirow{2}{*}{} & Song5 & 3.7 & 3.6 & 3.4 & 3.8 \\ 
		 \cmidrule{2-6}
		 & \textbf{Average} & \textbf{3.5} ±0.14 & 3.4 ±0.14 & 3.2 ±0.12 & 3.5 ±0.13 \\
        \midrule{}
        \multirow{6}{*}{\name{}} & Song1 & 3.5 & 3.1 & 3.3 & 3.5 \\
		 & Song2 & 3.4 & 3.7 & 3.5 & 4.0 \\
		 & Song3 & 3.2 & 3.6 & 3.3 & 3.6 \\
		 & Song4 & 2.9 & 3.0 & 3.1 & 3.5 \\
		 & Song5 & 3.4 & 3.6 & 3.2 & 3.9 \\ 
		 \cmidrule{2-6}
		 & \textbf{Average} & 3.3 ±0.15 & 3.4 ±0.15 & \textbf{3.3} ± 0.12 & \textbf{3.7} ±0.13 \\
        \bottomrule
	\end{tabular}
	}
	\caption{Qualitative evaluation results for \name{} w/o constraints and \name{}.}

	\label{tab:qualitative} 
\end{table}

\subsubsection{Qualitative Evaluation Results}
\label{sec:results}

To examine whether the proposed constraints improve the
singability and intelligibility, our qualitative evaluation compares \name{} with only length constraints to fully constrained \name{} (Table~\ref{tab:qualitative}) with \textit{syllable-to-syllable} assignment.
While the constraints significantly improve the intelligibility and slightly improve the singability (listening experience), these constraints make it harder for the original meaning to come through.
Overall, the annotators are satisfied with the translated songs by the
proposed baseline \name{}.
All aspects receive an average score around 3.5 out of 5.
These case studies and three translated songs by \name{} sung by an
amateur singer are available on
\href{https://gagast.github.io/posts/gagast}{https://gagast.github.io/posts/gagast}.

\section{Related Work}
\label{sec:rel}


\paragraph{Verse Generation and Translation.}
Generating verse text began through rule-based implementations~\cite{milic-70} and developed through the next forty years. \citet{manurung1999chart} design a chart system that generate strings that match a given stress
pattern. \citet{gervas2000wasp} build a forward reasoning rule-based system. \citet{manurung2004evolutionary} address poetry generation with stochastic search based on evolutionary algorithms.
\citet{oliveira2012poetryme} create a template-based platform that allows user to define features and create templates. 
\citet{he2012generating} adopt statistical machine translation models for Chinese poetry generation. 
\citet{yan2013poet} compose poetry based on generative summarization framework. 
\citet{zhang2014chinese}, \citet{wang2016chinese}, and \citet{hopkins2017automatically} adopt recurrent neural networks for poetry generation and incorporate rhythmic constraints. 
\citet{ghazvininejad2016generating,ghazvininejad2017hafez} represent rhythm and rhyme with finite-state machines.
Poetry translation using these frameworks and statistical machine translation thus offers elegant solutions: \citet{genzel2010poetic} intersect the finite state representation of the meter and rhyme scheme with the synchronous context-free grammar of the translation model under the phrase-based machine translation framework.
\citet{ghazvininejad-etal-2018-neural} apply the finite-state constraints to neural translation model.
However, these representations of the rhythmic and lexical constraints are not flexible enough to encode the real-valued representation of a \emph{song} as required for translation in tonal languages.

\paragraph{Constrained Text Generation.}
Most natural language generation tasks, including machine translation~\citep{bahdanau2014neural,vaswani2017attention,Hassan2018AchievingHP}, dialogue system~\citep{shang2015neural,Li2016APN,Wang2021TaskOrientedDS} and abstractive summarization~\citep{rush2015neural,Paulus2018ADR}, are free text generation. However, there is a need to generate text with constraints for special tasks~\citep{lakew2019controlling,li2020rigid,zou2021controllable}. ~\citet{hokamp-liu-2017-lexically,post-vilar-2018-fast,hu-etal-2019-improved} attempt to constrain the beam search with dictionary.
In the training procedure, \citet{li2020rigid} add format embedding. \citet{lakew2019controlling} introduce length tag. \citet{saboo-baumann-2019-integration} address length control via rescoring the results of beam search for machine translation under dubbing constraints. 


\paragraph{Lyrics Generation.}
As one of the most important tasks in automatic songwriting, lyrics generation has received more attention recently. \citet{sheng2020songmass}, \citet{lee2019icomposer} and \citet{chen2020melody} generate lyrics via pure data driven models without adding constraints based on expert knowledge.
\citet{oliveira2007tra} build a rule-based lyrics generation system to handle rhyme and rhythm with designed heuristics.
\citet{malmi2016dopelearning} address rap lyrics generation via information-retrieval approach and propose a rhyme-density measure. 
\citet{watanabe2018melody} add conditions in standard RNNLM with a featurized input melody for rhythmic alignment. 
\citet{ma2021ai} 
develop a SeqGAN-based lyrics generator to address various properties, such as rhythmic alignment, theme and genre. 
\citet{xue2021deeprapper} use transformer-based model to generate rap lyrics with a reverse order, address rhymes with vowel embeddings and add extra beat tokens for rthymic alignment.
We are the first paper that formally address the importance of aligning melody pitch with languages tones in lyrics generation for tonal languages. 
We introduce two vital qualities of songs, singability and intelligibility, and design three types of melody-lyric alignment scores to improve the two qualities.

\section{Conclusion}


This paper addresses automatic song translation~(\abr{ast}) for tonal languages and the unique challenge of aligning words' tones with melody. And we build the first English-Mandarin \abr{ast} system -- \name{}. Both objective and subjective evaluations demonstrate that \name{} successfully improves the singability and intelligibility of translated songs.

More constraints are left in the future work such as rhymes and style. We aim to build a systematic framework that address all constraints. 
With the help of newly developed singing voice synthesize tools such as X Studio,\footnote{https://singer.xiaoice.com} we can perform human evaluation with actual singing voice with a larger scale to provide more reliable analysis. 
Moreover, our system can also be applied in lyrics and song generation applications without translation input.

\section*{Ethical Considerations}

\name{} improves singability and intelligibility of the translated songs in Mandarin via constraining the decoding of a pretrained lyrics translation model. This methodology has limitations by imposing a direct trade-offs between the original objective and the constraints. In terms of negative impact or risks, the inaccurate translations may cause misunderstandings in applications like Musical Theatre.

This paper collects lyrics data that are publicly available and are parsed from the Web. We use these data for research purposes only. To prevent any abuse or piracy of these data, we chose the dataset license Attribution-NonCommercial-ShareAlike 4.0 International (CC BY-NC-SA 4.0).

\section*{Acknowledgements}

This work was initially supported by Alibaba \abr{damo} Academy
through Alibaba Innovative Research Program.
We thank the anonymous reviewers for helpful feedback on previous
versions of this work.
We thank Carol, Xin Xu and \abr{igp} for the helpful discussions about
music.
Finally, we thank the singer/songwriter Ayanga for his professional
discussion about the difficulties in musical translations, which was
the genesis of this paper.
Boyd-Graber and Guo's work is supported in part by a gift from Adobe.


\bibliography{bib/jbg,bib/journal-full,bib/fenfei}

\bibliographystyle{style/acl_natbib}

\clearpage

\section*{Appendix A}

\subsection*{A.1 Illustration of Tonal Alignment by Frequency} 

\begin{figure*}[t]
    \centering
    \includegraphics[width=0.9\linewidth]{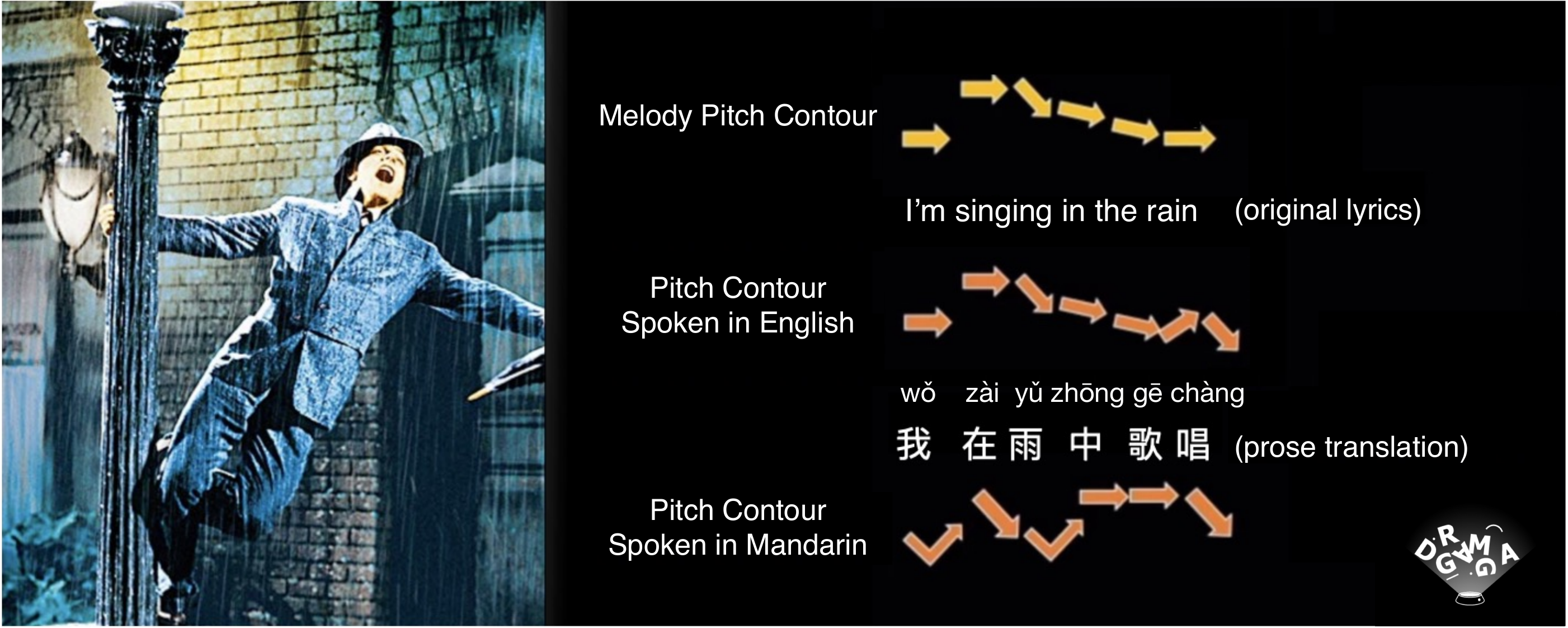}
    \caption{The pitch contour of the prose translation~(bottom line, in Mandarin) of lyrics do not match that of the original music~(upper line).  The directions showed in figure is estimated by the base frequency of spoken sound by text-to-speech tools. Such mismatch in pitch contour makes the sung lyrics sound unnatural and hard to understand.}
    \label{fig:illustration}
\end{figure*}

Translating songs into tonal languages faces a unique challenge,i.e., the tones of the translated lyrics should align with the music pitch for singability and intelligibility~(Section~\ref{sec:st_tonal}). Figure~\ref{fig:illustration} provides visual illustration of the main problem.

\begin{figure*}[t]
    \centering
    \includegraphics[width=0.9\linewidth]{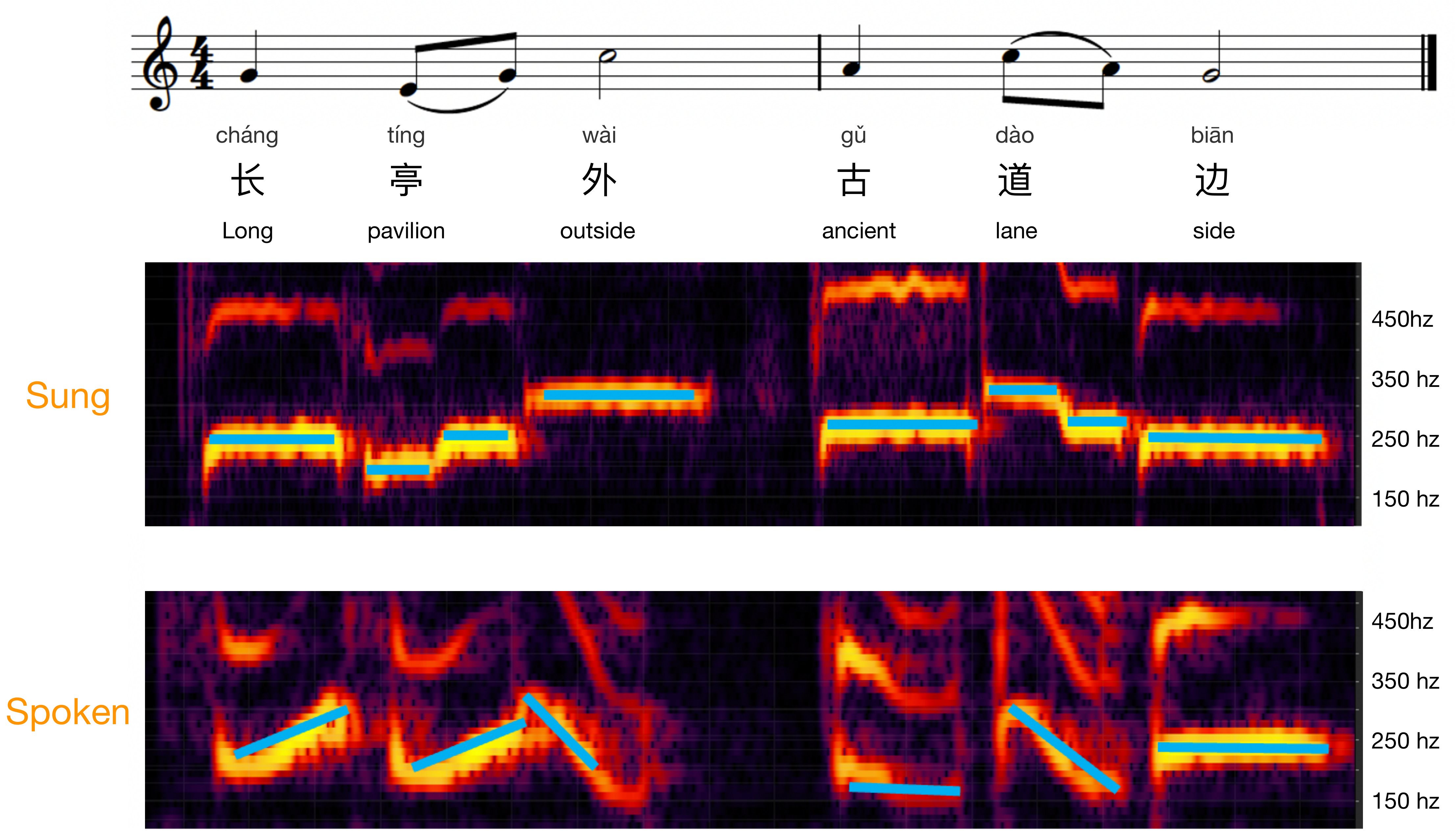}
    \caption{An example of a piece of a popular rewritten song in Mandarin ``Farewell (s\`ong bi\'e)''. The original music is from an American song ``Dreaming of Home and Mother''. We record the sung and spoken voice and plot the actual base frequency of the sound. We can see how the tone shape and overall tonal contour aligns with the sung voice~(by the music pitch).}
    \label{fig:sung_pitch}
\end{figure*}

To help researchers who speak non-tonal languages understand better how the tones of lyrics in tonal languages should align with the music/sung voice, we record both sung and spoken voice of a piece of lyrics from one of the most popular songs in Mandarin, transform the sound into the frequency space, and compare the shape of the sound with that of the music in Figure~\ref{fig:sung_pitch}. The original music of the chosen song is from an American song ``Dreaming of Home and Mother'', and was rewritten in Mandarin. Despite that this is not a translation task and do not have to convey the original meaning, we can see how the tonal contour of the lyrics in Mandarin align with that of music. 

\subsection*{A.2 Acceptable Pitch Transition Directions Table}

In Section~\ref{sec:st_mandarin}, we explain that in practice, the relative relationship of the pitch of the tones of the successive syllables/characters that belongs to the same word affect the most to the singability and intelligibility. And we summarize the acceptable transition directions in Figure~\ref{fig:tone_contour} under the assumption that only relative relationship of successive notes matters. It should be noted that we intend merely to provide a workable solution but not a perfect one. For example, the handle of the fourth notes of Mandarin is actually very tricky. It is a continuous fall with a large range~(see Figure~\ref{fig:tone_example}), therefore it doesn't represent one note. If it were to be represented by one note, it might represent the onset or offset part of the tone, and the falling trend is hinted by the pitch contour with proceeding and/or following note~\citep{zhuang1982peking,yu2008kunqu}.

\begin{table}[t]
  \small
	\centering
	\resizebox{\linewidth}{!}{
	\begin{tabular}{cc}
		\toprule
		 Parameter & Value \\
		\midrule
         encoder layer & 12   \\
         decoder layer & 12 \\
         max source position & 512  \\
         max target position & 512 \\
		 layernorm embedding & True  \\ 
		 criterion & label smoothed cross entropy \\
		 learning rate & 3e-4  \\ 
		 label smoothing & 0.2 \\
         min lr & 1e-9  \\
         lr scheduler & inverse sqrt \\
         warmup updates & 4000 \\
         warmup initial lr & 1e-7 \\
         optimizer & adam \\
         adam epsilon & 1e-6 \\
         adam betas & (0.9, 0.98) \\
         weight decay & 0.01 \\
         dropout & 0.1 \\
         attention dropout & 0.1 \\
         \midrule
         \multicolumn{2}{l}{\textit{text infilling}} \\
         mask rate & 0.3  \\
         poisson lambda & 3.5 \\
         replace length & 1 \\
        \bottomrule
	\end{tabular}}
	\caption{Pretraining hyper-parameters}
	\label{tab:params} 
\end{table}

\section*{Appendix B}

\subsection*{B.1 Training Details}

We pretrain our transformer-based model with reconstruction objective and corrupt our input sequence with \textbf{text infilling}~\citep{lewis-etal-2020-bart}. More detailed pretraining hyper-parameters can be found in Table~\ref{tab:params}.

\section*{Appendix C}

\subsection*{C.1 Human Evaluation Instruction}

In this paper, we conduct subjective evaluations by collecting annotations about the qualities of the translated songs from music school students~(Section~\ref{sec:human_eval}). 

\subsection*{C.2 Music Sheets}

\begin{figure*}[t]
    \centering
    \includegraphics[width=\linewidth]{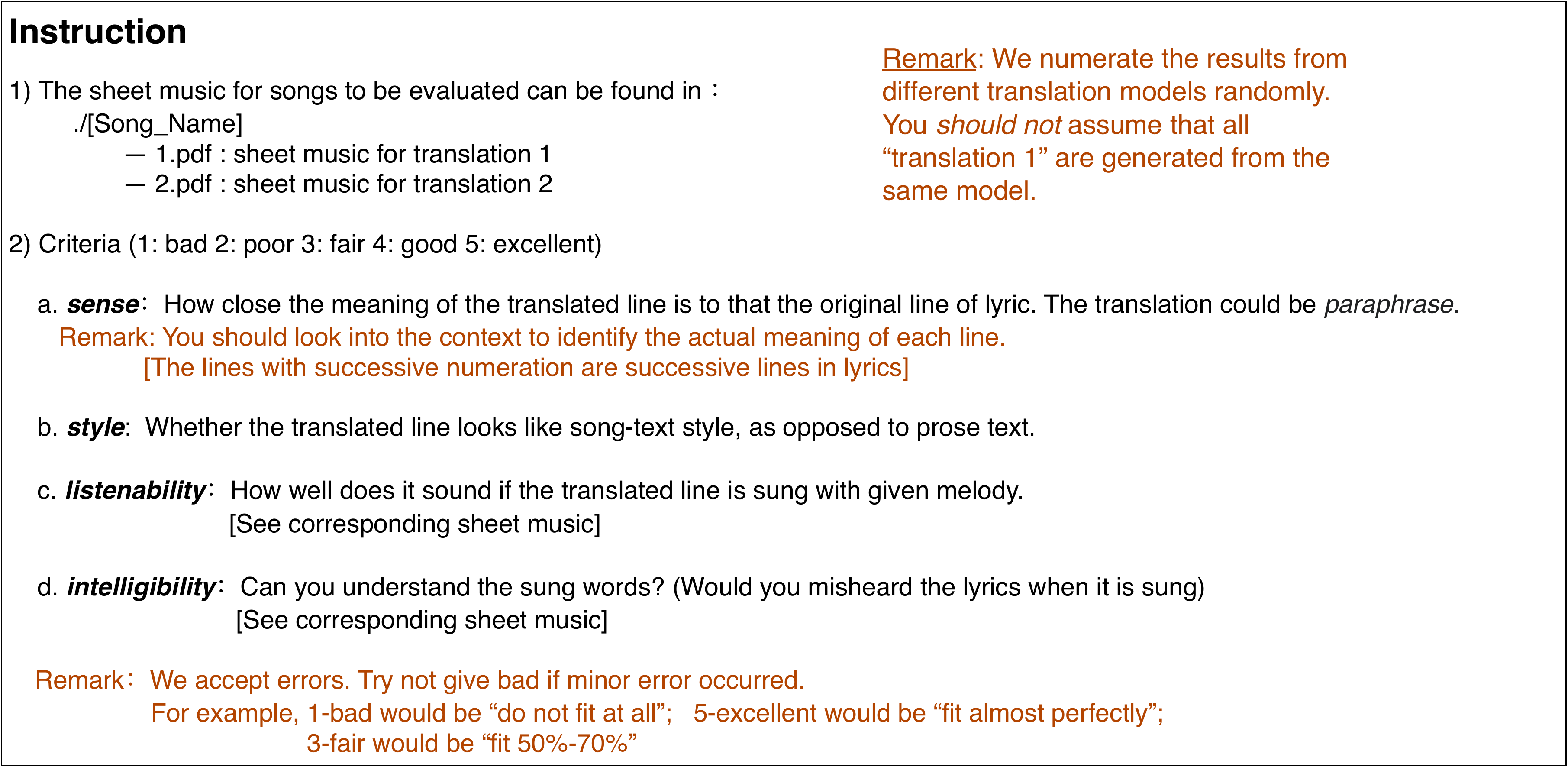}
    \caption{Instructions for human evaluation}
    \label{fig:instruction}
\end{figure*}

As describe in Section~\ref{sec:human_eval} and shown in the
instructions~(Figure~\ref{fig:instruction}), we distributes music
sheets of the translated songs to the annotators. All music sheets can
be found on \url{https://gagast.github.io/posts/human_eval}.

\end{document}


\maketitle
\section{Main Problem Illustration}
\begin{figure}[!thb]
    \centering
    \includegraphics[width=\textwidth]{2022_acl_song_translation/figures/illustration.pdf}
    \caption{An example where the pitch contour of the speech tone in Mandarin of the Chinese translation does not match the original melody; while the pitch contour (measured by TTS tools) of the original lyric in English align with the melody. As Mandarin is one of the tonal languages, a mismatch in tones makes the song sounds unnatural and the lyrics hard to understand.}
    \label{fig:illustration}
\end{figure}

\section{Datasets Details}
\subsection{Collection of Evaluation Dataset}
We tackle the lack of data in the training procedure by conduct AST in an unsupervised manner, however, for evaluation, we still need a small set of parallel data.
We collect the evaluation dataset in following steps:
\begin{itemize}
    \item Extract the (melody, source lyrics) paired dataset $\mathcal{A}$ from 100 public scores collected from the web. The source lyrics and melodies are aligned in the syllable-note level.
    \item Search the corresponding songs (source lyrics, target lyrics) $\mathcal{B}$  from the lyrics translation data which are also collected from the web~\footnote{https://lyricstranslate.com/}. 
    \item Considering that the data collected from the web is noisy, the source lyrics in $\mathcal{A}$ and $\mathcal{B}$ are not totally same. We align $\mathcal{A}$ and $\mathcal{B}$ by the source lyrics whose Levenshtein distance are less than a certain threshold.
\end{itemize}
 A subset of the collected evaluation dataset in our paper is in~\textbf{\textit{supplementary\_materials/data/eval\_set}}.
 
\subsection{General Translation Data}  We trained on the standard WMT 2014 English-Chinese dataset consisting of about 29.6 million sentence pairs. We attach a subset of WMT data in~\textbf{\textit{supplementary\_materials/data/wmt}}.
 
\subsection{Nonparallel Lyrics Data}
We collect large amount of lyrics of both Mandarin and English songs from the web, which contains about 12.4 million lines of lyrics for Mandarin and 109.5 million for English after removing the duplication. A subset of the collected nonparallel lyrics corpus in the supplementary material~\textbf{\textit{supplementary\_materials/data/nonparallel\_lyrics}}.
 
 \subsection{Lyrics Translation Data}
 We crawl a small set of lyrics translation data from the web~\footnote{https://lyricstranslate.com/}, which is Mandarin-English paired and contains 140 thousands pairs of lines. A subset of lyrics translation data is attached in~\textbf{\textit{supplementary\_materials/data/lyrics\_translation}}.
 
  \subsection{BPE codes}
  
  We preprocess all data with fastBPE~\citep{sennrich-etal-2016-neural} and a code size of 50,000.



\section{Subjective Evaluation}
\begin{table*}
  \small
	\centering
	\resizebox{0.7\linewidth}{!}{
	\begin{tabular}{cccccc}
		\toprule
		 Model & Song & \textit{sense} & \textit{style} & \textit{listenability} & \textit{intelligibility} \\
		\midrule
		 \multirow{2}{*}{} & Song1 & 3.4 & 3.0 & 3.2 & 3.4 \\
		 & Song2 & 3.6 & 3.9 & 3.5 & 3.8 \\
		 GagaST & Song3 & 3.7 & 3.6 & 3.4 & 3.5 \\
		 unconstrained & Song4 & 3.2 & 3.0 & 2.8 & 3.0 \\
		 \multirow{2}{*}{} & Song5 & 3.7 & 3.6 & 3.4 & 3.8 \\ 
		 \cmidrule{2-6}
		 & \textbf{Average} & \textbf{3.5} & 3.4 & 3.3 & 3.5 \\
        \midrule{}
        \multirow{6}{*}{GagaST} & Song1 & 3.5 & 3.1 & 3.3 & 3.5 \\
		 & Song2 & 3.4 & 3.7 & 3.5 & 4.0 \\
		 & Song3 & 3.2 & 3.6 & 3.3 & 3.6 \\
		 & Song4 & 2.9 & 3.0 & 3.1 & 3.5 \\
		 & Song5 & 3.4 & 3.6 & 3.2 & 3.9 \\ 
		 \cmidrule{2-6}
		 & \textbf{Average} & 3.3 & 3.4 & 3.3 & \textbf{3.7} \\
        \bottomrule
	\end{tabular}
	}
	\caption{Subjective evaluation results of GagaST-unconstrained and GagaST.}
	\label{tab:supp_subjective} 
\end{table*}
In this paper, as described in Section 4.2, we conduct human evaluations and compares our GagaST system with and without constraints (baseline). The instruction for annotators as shown in Figure~\ref{fig:instruction}. All annotators with domain knowledge are students in prestigious music school. 

In the full paper, we report the subjective evaluation results of 10 annotators over 20 samples. The 20 samples are randomly selected from the test set (total 713 samples) and not from the same song. 
In order to further explore if the context in a song can affect the subjective evaluation, we randomly select 5 songs from the test set and extract the first ten sentences of each song to construct the evaluation samples set.  We deliver those evaluation samples to 5 different annotators and all sheet music given to the annotators can be found in \textbf{\textit{supplementary\_materials/human\_evaluation/sheet\_music}}. An example is shown in Figure~\ref{fig:example} as well.

The results can be found in Table~\ref{tab:supp_subjective}. We can see that as our results on 20 samples, GagaST with constraints achieves better intelligibility for all five songs with a trade-off over sense. The increase on the intelligibility demonstrates that our designed rules for alignments are reasonable. 

However, our evaluation is done purely on sheet music, which is not the actual use case, i.e., an actual user listening to the song. In the future, we would conduct human evaluation with actual performed music.


\begin{figure}[t]
    \centering
    \includegraphics[width=\textwidth]{2021_neurips_lyrics_translation/figures/instructions.pdf}
    \caption{Instructions for human evaluation.}
    \label{fig:instruction}
\end{figure}

\begin{figure}[!thb]
    \centering
    \includegraphics[width=\textwidth, trim={0cm, 0cm, 0cm, 3.2cm}, clip=true]{2021_neurips_lyrics_translation/figures/example.pdf}
    \caption{A piece of sheet music with translated results from GagaST shown to annotators.}
    \label{fig:example}
\end{figure}


\section{Analyses}
\subsection{Length Tag}

\begin{wraptable}{r}{0.5\textwidth}
	\centering
	\vspace{-0.5cm}
	\resizebox{0.95\linewidth}{!}{
	\begin{tabular}{clcc}
		\toprule
		 Alignment & \multicolumn{1}{c}{\multirow{2}{*}{Model}} &  \multicolumn{2}{c}{Length}\\
		 Granularity & & \# of longer~$\downarrow$ & \# of shorter~$\downarrow$ \\
		\midrule
		 \multirow{2}{*}{note} & Baseline &  9 & 0  \\ 
		 & --w/o len tag &  302 & 274  \\ 
        \midrule
        \multirow{2}{*}{syllable}  & Baseline & 4 & 0  \\
        & --w/o len tag &  481 & 98 \\
        \bottomrule
	\end{tabular}
	}
	\caption{The effects of length tag}
	\label{tab:len_res}  
\end{wraptable}

In order to fit into the melody, we add length tag during pretraining to control the length of the translated lyrics. We provide a comparison in Table to show the effects of length tag. We can see that without length tag, the pretrained GagaST (unconstrained, baseline) is unable to generate lyrics that fit the length of given melody.

\subsection{Case Analyses}
\begin{figure}[!thb]
    \vspace{-0.2cm}
    \centering
    \includegraphics[width=\textwidth]{2021_neurips_lyrics_translation/figures/case_study_wmt.pdf}
    \caption{Case study -- We can see that even without parallel lyrics translation data, by cross-domain pretraining with nonparallel lyrics data, GagaST is able to translate lyrics that looks more like song text, and with more accurate meaning. Both results are decoded with same hyper-parameters for constraints and length tags (contour-0.5, shape-1.0, rest-3.0).}
    \label{fig:case_wmt}
\end{figure}

\begin{figure}[!thb]
\vspace{-0.2cm}
    \centering
    \includegraphics[width=\textwidth]{2021_neurips_lyrics_translation/figures/case_study_base.pdf}
    \caption{Case study -- We can see that with cross-domain pretraining, GagaST is able to generate lyrics that looks more like song text; and with constraints, GagaST is able to produce lyrics that are more "catchy". Both results are decoded with same hyper-parameters for constraints and length tags (contour-0.5, shape-1.0, rest-3.0).}
    \label{fig:case_constraints}
\end{figure}

\section{More Pretraining details}

For nonparallel lyrics data, we pretrain our model with reconstruction objective and corrupt our input sequence with \textbf{text infilling}~\citep{lewis-etal-2020-bart}. More detailed pretraining hyperparameters can be found in Table~\ref{tab:params}.

\begin{table*}[!thb]
  \small
	\centering
	\resizebox{0.8\linewidth}{!}{
	\begin{tabular}{cccc}
		\toprule
		 Parameter & Value & Parameter & Value \\
		\midrule
         encoder layer & 12 & decoder layer & 12 \\
         max source position & 512 & max target position & 512 \\
		 layernorm embedding & True & criterion & label smoothed cross entropy \\ 
		 learning rate & 3e-4 & label smoothing & 0.2  \\ 
         min lr & 1e-9 & lr scheduler & inverse sqrt \\
         warmup updates & 4000 & warmup initial lr & 1e-7 \\
         optimizer & adam & adam epsilon & 1e-6 \\
         adam betas & (0.9, 0.98) & weight decay & 0.01 \\
         dropout & 0.1 & attention dropout & 0.1 \\
         \midrule
         \multicolumn{4}{l}{\textit{text infilling}} \\
         mask rate & 0.3 & poisson lambda & 3.5 \\
         replace length & 1 & -- & -- \\
        \bottomrule
	\end{tabular}
	}
	\caption{Pretraining hyper-parameters}
	\label{tab:params} 
\end{table*}

\bibliography{bib/journal-full,bib/fenfei}

\bibliographystyle{style/acl_natbib}
